# WOVe: Incorporating Word Order in GloVe Word Embeddings

**Mohammed Ibrahim**

University of Arkansas, US, msibrahi@uark.edu,  https://orcid.org/0000-0001-6842-3745

**Susan Gauch**

University of Arkansas, US, sgauch@uark.edu,  https://orcid.org/0000-0001-5538-7343

**Tyler Gerth**

University of Arkansas, US, tdgerth@uark.edu

**Brandon Cox**

University of Arkansas, US, bscox@uark.edu

**Abstract**: Word vector representations open up new opportunities to extract useful information from unstructured text. Defining a word as a vector made it easy for the machine learning algorithms to understand a text and extract information from. Word vector representations have been used in many applications such word synonyms, word analogy, syntactic parsing, and many others. GloVe, based on word contexts and matrix vectorization, is an effective vector-learning algorithm. It improves on previous vector-learning algorithms. However, the GloVe model fails to explicitly consider the order in which words appear within their contexts. In this paper, multiple methods of incorporating word order in GloVe word embeddings are proposed. Experimental results show that our Word Order Vector (WOVe) word embeddings approach outperforms unmodified GloVe on the natural language tasks of analogy completion and word similarity. WOVe with direct concatenation slightly outperformed GloVe on the word similarity task, increasing average rank by 2%. However, it greatly improved on the GloVe baseline on a word analogy task, achieving an average 36.34% improvement in accuracy.

**Keywords:** Word embeddings, Vector learning, Attention mechanisms

## Introduction

Word embedding is the process of representing words as vectors of real numbers. These vectors can be used in many applications, such as document indexing (Hofmann, 2017), query expansion (Gauch & Chong, 1995), document classification (Lilleberg et al., 2015). Early approaches to vector representations for words created word-document co-occurrence matrices (Schütze & Pedersen, 1997). These vectors have high dimensions and can be very sparse, an issue addressed by Latent Semantic Analysis (LSA) (Deerwester et al., 1990) which performs single value decomposition on word-document matrix. (Finch & Chater, 1992) and (Gauch et al., 1999) developed word embeddings based on word-word co-occurrences. Given the computational limits of the time, they dealt with the high dimensionality (and sparseness) of the context matrix by limiting the dimensions of the vectors to a set of context-words.

Current approaches employ neural network-based models to learn word vectors. Mikolov and his team proposed the skip-gram and continuous-bag-of-words models using a single hidden layer of neural network (Mikolov, Chen, et al., 2013; Mikolov, Sutskever, et al., 2013). (Peters et al., 2018) presented a word embedding algorithm called ELMo that uses the Recurrent Neural Networks (RNN). ELMo considers the context of a word in a sentence. This mechanism helped producing different vectors for the same word and made ELMo different from the traditional word embedding algorithms. The most recent work in the field of word embeddings uses a new mechanism called the transformers. These transformers leverage a simple network architecture based on the concept of attention mechanisms. BERT (Devlin et al., 2019), GPT-2 (Radford et al., 2019), and CTRL (Keskar et al., 2019) are examples of recent algorithms that uses transformers to find best word embeddings.

In this paper, we improve the results of a traditional word embedding algorithm called the Global Vectors for word representations (GloVe). GloVe was proposed by (Pennington et al., 2014). This algorithm combines the advantages of matrix factorization of the word-word co-occurrence methods with local context windows. GloVe showed good results comparing to many traditional vector learning models. During our investigation of this algorithm, we found that the GloVe model fails to consider word position explicitly. When constructing the word-word co-occurrence matrix, increments are decreased based on their distance from the pivot word, but there is no other consideration taken. In this work, we introduce Word Order Vectors (WOVe) that extends the GloVe model by adding explicit representation of word position within the context window.

## Related Works

There has long been an interest in learning the meaning of words from the corpora in which they appear. Early approaches were based on the idea that a word can be defined from the documents in which they appear (Jones, 1971), and others assume that words can be determined from the contexts around them (Miller & Charles, 1991). Currently, word embedding algorithms have become a very active area of research, resulting in neural-based methods such as word2vec (Mikolov, Sutskever, et al., 2013)**,** GloVe (Pennington et al., 2014), ELMo (Peters et al., 2018). However, the new era of word embedding uses the transformers that showed promising results comparing to the neural-based models. These models have the ability to speed up the training process a way from the traditional word embedding training processes. (Radford et al., 2019) reported that their model, GTP-2, that uses the transformer mechanism, was able to provide state-of-the-art results with a zero-shot setting over 7 language modelling datasets.

For the GloVe algorithm, several approaches have been proposed to improve on its performance. Some approaches use explicit information to incorporate larger contexts by adding morphological information (Bojanowski et al., 2017; Gupta et al., 2019; Ibrahim et al., 2020; Nugaliyadde et al., 2019), and others did implicit updates incorporating different techniques (Mikolov, Sutskever, et al., 2013). (Bojanowski et al., 2017) used the n-gram technique to create vectors from the word characters rather than words themselves. Similarly, (Gupta et al.,

2019) used the kernel principal component analysis (KPCA) to enrich word embeddings with semantic and syntactic information. (Nugaliyadde et al., 2019) enhanced word vectors with word relationships. (Ibrahim et al., 2020) boosted the weights of target words on the co-occurrence matrix to increase the chance of finding their similar words on a healthcare corpus. In contrast, (Mikolov, Sutskever, et al., 2013) did not affect the word vectors directly. However, it improved the word embedding model by adding techniques such as using the negative sampling technique.

Our research is closer to (Bojanowski et al., 2017) and (Gupta et al., 2019) by adding more information to the vector. However, we improve GloVe algorithm by explicitly considering word position in the context of pivot word similar to (Gauch et al., 1999).

## Word Order Vectors

**GloVe**
Glove is one of the unsupervised learning algorithms that builds vectors to represent words in a corpus (Pennington et al., 2014). GloVe starts by constructing a global word-word co-occurrence matrix within a context window of a specific size, e.g., $\pm 10$. The value added for the word pair decreases as a function of distance, so that word pairs that are *d* words apart contribute *1/d* to the total count. GloVe uses a global log bilinear regression model to learn word vectors in a lower dimensionality space. It avoids the sparsity of the global co-occurrence matrix by running only on nonzero entries. GloVe has been shown to outperform other models on word analogy, word similarity and named entity recognition tasks.

**Context Word Vectors**
(Gauch & Futrelle, 1994) developed a positional context vector approach that they used for disambiguation and to classify words into their parts of speech. This work was adapted to learn word similarities that were shown to effectively identify related words for automatic query expansion (Gauch et al., 1999). A major feature of this work was the use of a separate co-occurrence counts for context word observations at each position within the context window. By incorporating word position, the resulting words were ordered by syntactic as well as semantic similarity.

**Word Order Vectors (WOVe)**
Our goal is to combine the previous two approaches by enhancing GloVe to explicitly incorporate word positions. We modified GloVe to create a word-word co-occurrence matrix for a given position within a context window, e.g., $\pm 2$. We then had GloVe create *positional word vectors* for each position within the context window around the pivot word, *w*. We developed and evaluated three different methods of using the vectors GloVe as described below.

<u>Direct Concatenation</u> In this method, the dimensionality of the vectors is the same for each positional word vector and the regular GloVe word vectors. The positional word vectors are concatenated to create the final word vectors. Thus, if the size of the GloVe vector is *k* and the context window size is *c*, the resulting word vector would have dimensionality *ck*.

Reduced Concatenation Because we create *c* word co-occurrence matrices, each is based on only *1/c* word appearances rather than the c word appearances used by GloVe in its single matrix. This method addresses this data sparsity and the increased vector size that the direct concatenation method produces by creating positional word vectors of length *k/c*. Thus, after concatenation, the final word vectors are length *k* as is the case with unmodified GloVe.

Weighted Direct Concatenation The concatenation methods do not take into account the possibility that words closer to the pivot word are more likely to be related to it. Similar to GloVe, the weighted concatenation method decreases the contribution of each positional word vector proportionally by its distance from the pivot word, *w*. Thus, each positional word vector has its weight multiplied by *1/d* where d is that position's distance from *w*, prior to concatenation. Because our early results showed that direct concatenation greatly outperformed reduced concatenation, we only evaluated proportional weightings for that algorithm.

## Evaluation

### Experiment Settings

The methods proposed in this paper were evaluated using 8.7 GB of text randomly selected from a 2019 corpus of Wikipedia text (*Enwiki Dump Progress on 20190920*, n.d.). The dataset was cleaned by removing XML tags, punctuation, and whitespace characters other than spaces between words. After that, the text is tokenized and downcased. The resulting dataset contained a total of 1,372,327,637 tokens.

We used the basic GloVe architecture reported in (Pennington et al., 2014) as the baseline against which we compared our algorithms. We did not compare WOVe with other word embedding algorithms such as Word2Vec (Mikolov, Sutskever, et al., 2013), vLBL (Mnih & Kavukcuoglu, 2013), and others because (Pennington et al., 2014) already reported that GloVe outperformed many of these algorithms. We also did not compare with the current state-of-the-art embedding algorithms such as BERT (Devlin et al., 2019), and ELMO (Peters et al., 2018) because these algorithms are bidirectional algorithms while WOVe is a single word embedding algorithm.

We evaluated our system on two tasks: (1) word analogy task and (2) word similarity task. The word analogy task consists of 14 sets of analogies selected from the dataset that comes with the GloVe code (*GloVe: Global Vectors for Word Representation*, n.d.). These were comprised of 19,544 analogies that should be answerable given a large corpus of Wikipedia data. They contain analogies such as "Chicago is to Illinois as Dallas is to Texas," and "Boy is to girl as father is to mother,". These analogies evaluate both syntactic and semantic of the words in the corpus. Equation 3 used to assess analogies:

$$A - B + C = D \qquad (2)$$

where *A*, *B*, *C*, and *D* are vectors that GloVe and its improved versions should create. We evaluated the algorithms using accuracy, i.e., the percent of the time that the word embedding predicted the correct answer (*D* in equation 2) given *A*, *B*, and *C*.

For the similarity task we collected five different datasets that contained human generated synonyms for words ((Miller & Charles, 1991) MC, (Rubenstein & Goodenough, 1965) RG, WordSim353 (Agirre et al., 2009), Stanford's Contextual Word Similarities SCWS (Huang et al., 2012), and Stanford's Rare Word Similarity Dataset RW (Luong et al., 2013). These contained 1+ synonyms for each target word. We measured the average rank of the synonyms that algorithms find within range of ten most similar words to every test word.

In the baselines and all the proposed methods, words with frequency less than five were excluded. The word embeddings vector size was fixed at 100 dimensions and the context window size varied from 2 to 10, using positions $\pm 1$ to $\pm 5$. Other GloVe parameters were set to the best setting reported in (Pennington et al., 2014).

**Results and Discussion**

Table 1 shows the number of correct analogies that GloVe and each of the positional enhancements were able to predict and the associated relative improvements for the WOVe methods versus the GloVe baseline. From these results, we see that direct concatenation, the best WOVe algorithm, improves GloVe's baseline by an average 36% across all tested context window sizes. This method also finds the most correct analogies, 8892, with a context size of $\pm 4$., although the context size of $\pm 5$ with 8891 correct performs almost identically.

Table 1: #Correct predictions and relative improvement for GloVe and WOVe with context windows from $\pm 1$ to $\pm 5$.

| Algorithm | 1 | 2 | 3 | 4 | 5 | Average |
|---|---|---|---|---|---|---|
| GloVe | 2709 | 5314 | 6532 | 7096 | 7089 | 5748 |
| Direct Con | 4273 | 7516 | 8517 | **8892** | 8891 | 7617 |
| RelImprov | 57.73% | 41.43% | 30.38% | 25.31% | 25.42% | 36% |
| Reduced Con | 1559 | 577 | 103 | 22 | 3 | 452 |
| RelImprov | -42.45% | -89.14% | -98.4% | -99.69% | -99.9% | -85.93% |
| WtDirect Con | 4336 | 6383 | 6273 | 6978 | 7238 | 6241 |
| RelImprov | 60.05% | 20.11% | -3.96% | -1.66% | 2.10% | 15.32% |

Weighting the word contributions by distance also improves the accuracy versus the baseline, but by a smaller amount. We suspect that this is because the positions are accounted for in the separate vectors so adding weights has little effect on the outcome. Finally, the reduced concatenation model underperforms GloVe badly, decreasing accuracy by an average 70.36% relative, no doubt because each of the position vectors is mapped into fewer and fewer dimensions as the context window sizes increase.

Table 2 shows the results of comparing our best WOVe algorithm to GloVe on the word synonym task. We compared the rank of the true synonym, from 0..9, in the list of 10 most similar words generated by each approach. If the human-generated synonym did not appear in the algorithm-generated list, it was assigned a 10 to denote that it was not found. In general, WOVe algorithm improved GloVe's baseline by an average 2% across all similarity datasets using a window size of ±4. The lowest rank (best results) occurred using wordsim353, with rank of 9.12, a 4% improvement. On average, WOVe found eight more synonyms than GloVe, a 34% improvement. The largest difference between the two algorithms in the number of synonyms discovered was 17 synonyms for the RW dataset with 46% improvement.

Table 2: Average rank and the number of synonyms that GloVe and WOVe found for different similarity datasets with window size ±4.

| Algorithm \ Datasets | MC | RG | WordSim-353 | SCWS | RW | | Average |
|---|---|---|---|---|---|---|---|
| **GloVe** | | | | | | | |
| Avg rank | 9.40 | 9.62 | 9.47 | 8.72 | 9.9 | | 9.424 |
| Synonyms | 4 | 4 | 24 | 316 | 37 | | 77 |
| **Direct concatenation** | | | | | | | |
| Avg rank | 9.14 (3%) | 9.47 (2%) | 9.12 (4%) | 8.64 (1%) | 9.8 (0.03%) | | 9.248 (2%) |
| Synonyms | 5 (25%) | 7 (75%) | 29 (21%) | 330 (4.5%) | 54 (46%) | | 85 (34%) |

## Conclusion and Future work

The main goal of this research was to incorporate word order into GloVe word embedding algorithm. We present three word order vector (WOVe) methods: direct concatenation, reduced concatenation, and weighted direct concatenation. We found that WOVe with direct concatenation slightly outperformed GloVe on the word similarity task, increasing average rank by 2%. However, it greatly improved on the GloVe baseline on a word analogy task, achieving an average 36.34% improvement in accuracy. For future work, we suggest fine-tuning of the WOVe to find not only word embeddings but also sentence embeddings and compare it with state-of-the-art sentence embedding algorithms.

# References

Agirre, E., Alfonseca, E., Hall, K., Kravalova, J., Paşca, M., & Soroa, A. (2009). A Study on Similarity and Relatedness Using Distributional and WordNet-based Approaches. *Proceedings of Human Language Technologies: The 2009 Annual Conference of the North American Chapter of the Association for Computational Linguistics*, 19–27. https://www.aclweb.org/anthology/N09-1003

Bojanowski, P., Grave, E., Joulin, A., & Mikolov, T. (2017). Enriching word vectors with subword information. *Transactions of the Association for Computational Linguistics*, *5*, 135–146.

Deerwester, S., Dumais, S. T., Furnas, G. W., Landauer, T. K., & Harshman, R. (1990). Indexing by latent semantic analysis. *Journal of the American Society for Information Science*, *41*(6), 391–407.

Devlin, J., Chang, M.-W., Lee, K., & Toutanova, K. (2019). BERT: Pre-training of Deep Bidirectional Transformers for Language Understanding. *ArXiv:1810.04805 [Cs]*. http://arxiv.org/abs/1810.04805

*Enwiki dump progress on 20190920*. (n.d.). Retrieved December 6, 2019, from https://dumps.wikimedia.org/enwiki/20190920/

Finch, S., & Chater, N. (1992). Bootstrapping syntactic categories using statistical methods. *Background and Experiments in Machine Learning of Natural Language*, *229*, 235.

Gauch, S., & Chong, M. K. (1995). *Automatic Word Similarity Detection for TREC 4 Query Expansion*. 10.

Gauch, S., & Futrelle, R. P. (1994). Experiments in automatic word class and word sense identification for information retrieval. *Proceedings of the Third Annual Symposium on Document Analysis and Information Retrieval*, 425–434.

Gauch, S., Wang, J., & Rachakonda, S. M. (1999). A corpus analysis approach for automatic query expansion and its extension to multiple databases. *ACM Transactions on Information Systems (TOIS)*, *17*(3), 250–269.

*GloVe: Global Vectors for Word Representation*. (n.d.). Retrieved February 8, 2021, from https://nlp.stanford.edu/projects/glove/

Gupta, V., Giesselbach, S., Rüping, S., & Bauckhage, C. (2019). Improving Word Embeddings Using Kernel PCA. *Proceedings of the 4th Workshop on Representation Learning for NLP (RepL4NLP-2019)*, 200–208. https://doi.org/10.18653/v1/W19-4323

Hofmann, T. (2017). Probabilistic Latent Semantic Indexing. *SIGIR Forum*, *51*(2), 211–218. https://doi.org/10.1145/3130348.3130370

Huang, E., Socher, R., Manning, C., & Ng, A. (2012). Improving Word Representations via Global Context and Multiple Word Prototypes. *Proceedings of the 50th Annual Meeting of the Association for Computational Linguistics (Volume 1: Long Papers)*, 873–882. https://www.aclweb.org/anthology/P12-1092

Ibrahim, M., Gauch, S., Salman, O., & Alqahatani, M. (2020). Enriching Consumer Health Vocabulary Using Enhanced GloVe Word Embedding. *ArXiv:2004.00150 [Cs, Stat]*. http://arxiv.org/abs/2004.00150

Jones, S. (1971). Automatic Keyword Classification for Information Retrieval. *Butterworths, London, UK*.

Keskar, N. S., McCann, B., Varshney, L. R., Xiong, C., & Socher, R. (2019). CTRL: A Conditional Transformer Language Model for Controllable Generation. *ArXiv:1909.05858 [Cs]*. http://arxiv.org/abs/1909.05858

Lilleberg, J., Zhu, Y., & Zhang, Y. (2015). Support vector machines and Word2vec for text classification with semantic features. *2015 IEEE 14th International Conference on Cognitive Informatics Cognitive Computing (ICCI*CC)*, 136–140. https://doi.org/10.1109/ICCI-CC.2015.7259377

Luong, T., Socher, R., & Manning, C. (2013). Better Word Representations with Recursive Neural Networks for Morphology. *Proceedings of the Seventeenth Conference on Computational Natural Language Learning*, 104–113. https://www.aclweb.org/anthology/W13-3512

Mikolov, T., Chen, K., Corrado, G., & Dean, J. (2013). Efficient estimation of word representations in vector space. *ArXiv Preprint ArXiv:1301.3781*.

Mikolov, T., Sutskever, I., Chen, K., Corrado, G. S., & Dean, J. (2013). Distributed representations of words and phrases and their compositionality. *Advances in Neural Information Processing Systems*, 3111–3119.

Miller, G. A., & Charles, W. G. (1991). Contextual correlates of semantic similarity. *Language and Cognitive Processes*, *6*(1), 1–28.

Mnih, A., & Kavukcuoglu, K. (2013). Learning word embeddings efficiently with noise-contrastive estimation. In C. J. C. Burges, L. Bottou, M. Welling, Z. Ghahramani, & K. Q. Weinberger (Eds.), *Advances in Neural Information Processing Systems 26* (pp. 2265–2273). Curran Associates, Inc. http://papers.nips.cc/paper/5165-learning-word-embeddings-efficiently-with-noise-contrastive-estimation.pdf

Nugaliyadde, A., Wong, K. W., Sohel, F., & Xie, H. (2019). Enhancing semantic word representations by embedding deep word relationships. *Proceedings of the 2019 11th International Conference on Computer and Automation Engineering*, 82–87.

Pennington, J., Socher, R., & Manning, C. (2014). Glove: Global vectors for word representation. *Proceedings of the 2014 Conference on Empirical Methods in Natural Language Processing (EMNLP)*, 1532–1543.

Peters, M. E., Neumann, M., Iyyer, M., Gardner, M., Clark, C., Lee, K., & Zettlemoyer, L. (2018). Deep contextualized word representations. *ArXiv:1802.05365 [Cs]*. http://arxiv.org/abs/1802.05365

Radford, A., Wu, J., Child, R., Luan, D., Amodei, D., & Sutskever, I. (2019). *Language Models are Unsupervised Multitask Learners*. 24.

Rubenstein, H., & Goodenough, J. B. (1965). Contextual correlates of synonymy. *Communications of the ACM*, *8*(10), 627–633. https://doi.org/10.1145/365628.365657

Schütze, H., & Pedersen, J. O. (1997). A cooccurrence-based thesaurus and two applications to information retrieval. *Information Processing & Management*, *33*(3), 307–318.